\documentclass[11pt]{article}

\usepackage[final]{acl}
\usepackage{placeins}

\usepackage{times}
\usepackage{latexsym}

\usepackage[T1]{fontenc}

\usepackage[utf8]{inputenc}

\usepackage{microtype}

\usepackage{inconsolata}

\usepackage{graphicx}
\usepackage{array}

\usepackage{booktabs}
\usepackage{multirow}
\usepackage{colortbl}
\usepackage{graphicx}
\usepackage{float} 
\usepackage{amsmath}
\usepackage{xcolor}
\usepackage[dvipsnames]{xcolor}
\usepackage{flafter} 
\usepackage{hyperref}

\definecolor{openai}{RGB}{200, 230, 245}    
\definecolor{metaai}{RGB}{220, 235, 220}    
\definecolor{omni}{RGB}{230, 220, 245}      
\definecolor{ourmodels}{RGB}{240, 220, 200} 

\title{TeleASR-Amharic: A Benchmark for Robust Amharic ASR in Realistic Telecommunication Conditions}

\author{\normalsize Dagnachew Mekonnen Marilign$^{1}$,  Enrique Molero$^{2}$,  Abrham Haile$^{3}$, Louis Powell$^{2}$,   \\ 
\textbf{\normalsize  Israel Abebe Azime$^{4}$ } \\
\footnotesize
$^1$ HiLCoE School of Computer Science and Technology, $^2$ GSMA ,  $^3$ gheero, 
\\
\footnotesize 
$^4$Saarland University 
}

\begin{document}
\maketitle

\begin{abstract}
The real-world usability of ASR models remains challenging, particularly for low-resource languages, where existing speech resources are often collected in clean studio environments. In this work, we introduce TeleASR-Amharic, a benchmark for evaluating Amharic ASR under various acoustic and telecommunication conditions. We analyze the major degradations observed in real telecommunications recordings and develop a benchmark that covers multiple channel conditions, environmental disturbances, and telecom-domain variations. Our results show that telecommunication degradation can substantially reduce recognition performance, making even high-performing ASR models impractical under severe conditions. For example, the WER of the best-performing model’s increases from 0.23 to 0.79 under severe telecommunication degradation, corresponding to a relative increase of 243\%   in recognition errors. We further investigate the effects of different noise and channel distortions and show that scaling training data and applying carefully designed telecommunication-aware augmentation substantially improve robustness. In particular, our best training configuration achieves a 23.6\% relative WER reduction under extreme noise conditions. Together with the release of our telecommunications-domain data, these findings provide a reproducible benchmark for measuring progress in Amharic ASR while highlighting open challenges in developing robust speech technologies for real-world deployment.

\end{abstract}

\section{Introduction}

\begin{figure}
    \centering
\includegraphics[width=0.8\linewidth]{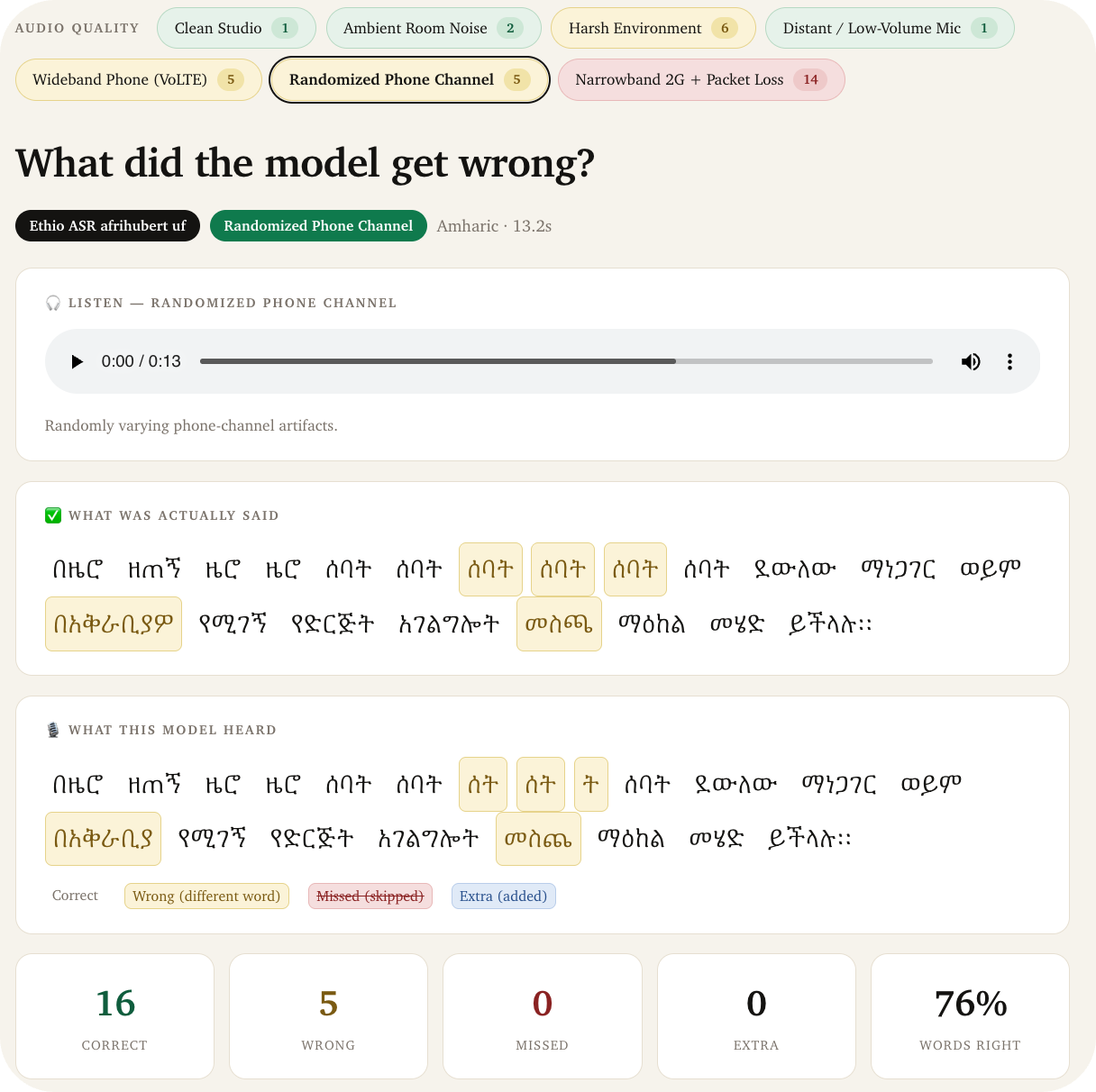}
    \caption{Examples of problematic ASR transcriptions in the telecommunications domain. The analysis shows that models frequently miss numeric information and that transcription performance varies substantially across different audio-quality conditions.}
    \label{fig:placeholder}
\end{figure}

Automatic Speech Recognition (ASR) has advanced rapidly in recent years, driven by significant developments in supervised learning, self-supervised representation learning, and multilingual modeling \cite{yadav-sitaram-2022-survey,9893562,AHLAWAT2025201,lugo-vielzeuf-2024-towards}. These advances, together with the emergence of large-scale speech models, have substantially expanded the range of languages, speakers, and acoustic conditions that contemporary ASR systems can handle. However, strong performance under clean or controlled speech does not necessarily translate into reliable recognition in real-world communication environments~\cite{Pranida2025ASRUN}. 

Unlike conventional clean-speech benchmarks, telecommunication speech can exhibit combinations of bandwidth limitations, sampling-rate reductions, dynamic-range compressions, nonlinear distortions, electrical interferences, background noises, and temporary signal losses~\cite{paul2026buildingevaluatingsyntheticbengali}. Such distortions may occur at different stages of the communication process and can vary substantially in terms of severity. Consequently, evaluating ASR systems only on clean or naturally recorded speech provides limited information about their robustness to the conditions encountered in real-world communication channels \cite{imai-etal-2025-evaluating, pranida-etal-2025-asr}. Therefore, a systematic evaluation framework that isolates and controls these factors is necessary to understand how ASR systems respond to progressively degraded speech.

Publicly available Amharic speech and transcription resources are substantially smaller than those available for high-resource languages \cite{nigatu-aldarmaki-2025-exploring, clearglobal2026lessons}. Recent efforts have expanded Amharic speech resources and demonstrated the feasibility of constructing larger corpora \cite{diack2026waxallargescalemultilingualafrican,yeneneh2025amharic, whisper_2026}. Self-supervised and multilingual models can further reduce the dependence on large amounts of transcribed data by transferring representations learned from multilingual speech \cite{baevski2020wav2vec, pratap2023scaling}. Nevertheless, increasing linguistic coverage does not necessarily resolve the acoustic domain mismatch. A model may adequately represent Amharic linguistic patterns while remaining sensitive to the signal degradation introduced by microphones, mobile devices, and communication networks.

To investigate this problem, we analyze the acoustic characteristics of recordings of Ethio Telecom and identify signal degradations that distinguish telecommunication speech from clean speech. Based on these observations, we constructed \emph{TeleASR-Amharic}, a benchmark to evaluate the robustness of ASR systems under controlled telecommunication-oriented acoustic conditions. Rather than representing degraded speech as a single evaluation category, the benchmark models a range of environmental, hardware, and communication-channel distortions at different severity levels. These include bandwidth limitation, sampling-rate reduction, dynamic-range compression, nonlinear distortion, background noise, power-line interference, and temporal signal dropouts. To deal with these problems, we ask \textbf{RQ1:} How robust are Amharic-specific, multilingual, and general-purpose ASR models to telecommunication-specific acoustic degradation? Next, we explore~\textbf{RQ2:} how ASR performance changes as the severity and combination of telecommunication degradations increase? to deal with this problems of domain mismatch and noise robustness we \textbf{RQ3:} explore the effect of data scaling and targeted augmentation to improve the robustness of Amharic ASR models under telecommunication conditions?



To answer these questions, we made the following contributions:

\begin{itemize}
    \item We introduce \textbf{TeleASR-Amharic}, a benchmark for evaluating the robustness of Amharic ASR systems under controlled, telecommunication oriented acoustic conditions. Data relesed under a CC BY 4.0 license with free access\footnote{\url{GSMA/TeleASR-Amharic}}.
    \item We provide a comprehensive evaluation of Amharic-specific, multilingual, and general-purpose ASR models across clean and progressively degraded conditions, enabling direct comparison of their robustness to telecommunication effects.

    \item We investigate the contribution of data scaling and targeted augmentation to robust Amharic ASR and analyze the behavior of the model.
\end{itemize}

\begin{figure*}[ht]
    \centering
\includegraphics[width=\linewidth]{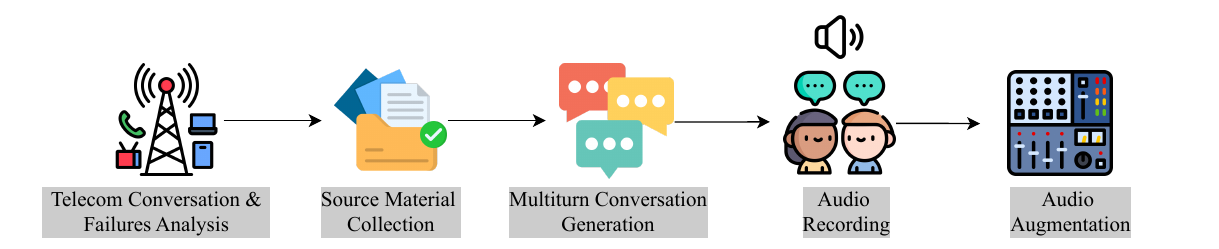}
    \caption{The process of creating the ASR robustness benchmark for the telecommunications domain. Each step is discussed in detail in Section~\ref{data}.}
    \label{fig:data-pipeline}
\end{figure*}

\section{Related Work}

\paragraph{Amharic ASR and Low-Resource Challenge}

Amharic is one of Ethiopia’s major languages and is widely used across public services, education, media, and digital applications. However, it remains under-resourced, with limited transcribed speech and substantial variation across speakers, dialects, devices, acoustic environments, and speaking styles \cite{nigatu-aldarmaki-2025-exploring, imam-etal-2025-automatic, bandarupalli-etal-2025-efficient, bartelds-etal-2023-making, liu-etal-2024-advancing, dhasmana-etal-2026-dialect}. This creates both linguistic and acoustic coverage gaps, as existing corpora often lack the variability encountered in real-world deployment. Therefore, we evaluate Amharic ASR beyond clean speech, focusing on robustness to realistic telecommunication conditions and domain-specific content, including numbers, terminology, and expressions used in customer-support interactions.

\paragraph{Telecommunication Speech as a Distinct ASR Domain}

Telecommunication speech represents a distinct ASR domain because transmission channels introduce systematic acoustic transformations beyond additive noise. Narrow-band telephony typically operates at 300–3400 Hz, limiting phonetic information and creating a domain shift that can degrade recognition performance \cite{itutp310, itutP341, itutg722}. Sampling rates impose further spectral constraints, with 8 kHz sampling limiting the representable spectrum to 4 kHz \cite{itutg711, itutg7111_2012}. Studies show that bandwidth mismatch between training and evaluation speech can substantially reduce ASR performance \. We therefore explicitly evaluate Amharic ASR robustness under realistic telecommunication channel conditions and transmission-induced degradation \cite{10.1007/s11042-020-10073-7}.

\paragraph{Robust ASR and the Role of Data Augmentation}

Robust ASR has traditionally addressed acoustic mismatch through normalization, noise compensation, adaptation, enhancement, and multi-condition training \cite{shah2025revisiting, wang2026bridging}. Data augmentation provides a complementary approach by exposing models to diverse acoustic conditions and reducing sensitivity to nuisance variability \cite{mengke2026can, 10.1016/j.compeleceng.2025.110851, venkateswaran-liu-2024-looking}. Established methods such as SpecAugment and waveform-level perturbation improve robustness, whereas recent work emphasizes acoustic-aware and deployment-specific augmentation \cite{park2019specaugment, ko2015audio, liu2025pretrainingrobustasrfoundation, zhang2021sensing, hao-etal-2026-investigating}. For telecommunications, augmentation should capture interacting effects such as bandwidth limitations, resampling, compression, nonlinear distortion, noise, and temporal signal loss \cite{wang2026augcodeclowbitratedisentangledneural}. Randomized channel augmentation can further capture variability by sampling distortion types and severities rather than relying on a fixed channel profile \cite{lin-etal-2025-channel, bartelds-etal-2023-making}. We apply this principle to Amharic through telecommunication-aware augmentation based on observed channel degradation and evaluate robustness under both controlled and randomized conditions.

\paragraph{Synthetic Audio Data Creation}

Collecting and manually transcribing speech across diverse deployment conditions is costly, making synthetic audio generation a practical way to expand acoustic coverage without new linguistic annotations \cite{casanova2023asr, radford2023robust, bartelds-etal-2023-making, hao-etal-2026-investigating}. 
This is particularly useful for low-resource ASR, where synthetic augmentation can increase training data and acoustic diversity through realistic channel, noise, and codec variations \cite{derenzi-etal-2025-synthetic, moslem-2024-leveraging, li-etal-2025-kits, joshi-singh-2022-simple, casanova2023asr, radford2023robust}. In this work, we generate telecommunication-degraded Amharic speech from existing transcribed utterances using controlled channel transformations designed to reflect realistic Ethiopian telecommunication conditions.

\section{TeleASR-Amharic Dataset}
\label{data}

Below, we describe each step involved in creating our telecom robustness benchmark, as illustrated in Figure~\ref{fig:data-pipeline}. Although synthetic augmentation may not fully capture real-world scenarios, it provides a controlled setting for studying model robustness. By introducing specific types of noise while keeping the domain and speaker identity constant, we can isolate the effects of individual perturbations and investigate whether observed performance degradation is attributable to the introduced noise rather than changes in domain or speaker characteristics.


\subsection{EthioTelecom Conversation \& Failures Analysis}

TeleASR-Amharic was motivated by the lack of robust ASR evaluation benchmarks that reflect real-world telecommunications customer-support scenarios. We first evaluated several ASR models on internal Ethio Telecom recordings and observed substantial recognition errors, particularly under conditions representative of real-world deployments. Our analysis identified two key challenges.

First, acoustic and channel degradation: Ethio Telecom recordings exhibit the characteristics of narrowband telephony, low-bitrate speech codecs, and heterogeneous mobile network conditions. These artifacts alter the spectral and temporal characteristics of speech, thereby creating additional challenges for automatic speech recognition.

Second, numerical information: Telecommunications customer support interactions frequently involve numbers, including phone numbers, account identifiers, balances, dates, and transaction amounts. We observed that ASR models struggled to represent such information consistently, often mixing written number words with numerical digits. This inconsistency is particularly problematic for downstream telecommunications applications, in which the accurate transcription of numerical information is essential.

Although these observations were derived from real Ethio Telecom recordings, the recordings cannot be used or released as a benchmark because they contain sensitive information, including names, phone numbers, account details, and payment-related discussions. Therefore, we developed TeleASR-Amharic using synthetic speech that preserves the relevant linguistic and acoustic characteristics of telecommunications interactions without exposing real individuals or their sensitive information.

\subsection{Source Material Collection}
\begin{table}[ht]
\centering
\scalebox{0.9}{
\begin{tabular}{lrrr}
\toprule
\textbf{Gender} & \textbf{Utterances} & \textbf{Duration (h)} & \textbf{Share \%} \\
\midrule
Female & 9,967 & 18.63 & 60.8 \\
Male   & 5,998 & 11.99 & 39.2 \\
\midrule
\textbf{All} & \textbf{15,965} & \textbf{30.63} & \\
\bottomrule
\end{tabular}
}
\caption{Speech duration statistics by gender for the TeleASR-Amharic dataset.}
\label{tab:gender-hours}
\end{table}

We collected real-life transcribed conversations, SMS support use cases, customer support FAQs, and agent training materials to capture realistic domain vocabulary and typical customer scenarios.

\subsection{Multiturn Conversation Generation}

Using the collected source materials as grounding documents, we generated 10-turn user–assistant conversations covering numbers, culturally specific names, and common telecommunications problems. The generated conversations were reviewed by Ethio Telecom experts and regenerated to address the identified issues, including inconsistencies in greetings and required closing statements specified by Ethio Telecom protocols.

\subsection{Audio Recording}

Eight paid contributors were selected from an existing contributor pool based on their record of approved speech-data submissions. Their speech varieties were identified through a dialect-identification task and verified by human reviewers. The selected speakers represented the Addis Ababa and Shewa varieties of Amharic. For each scripted dialogue, contributors were paired, with one acting as a telecom customer and the other as a customer-service agent. Recording tasks were distributed through the gheero crowd-sourcing platform\footnote{\url{https://gheero.et/}}. Each contributor used a smartphone to record the assigned utterances in the same studio, with customer and agent turns stored as separate audio files. Before recording, all contributors provided informed consent, and direct identifiers were removed from the associated metadata. The detailed dataset statistics of the final data are shown in Table~\ref{tab:gender-hours}.

\paragraph{Quality Control.}

A human reviewer listened to every submitted recording and compared it with the corresponding script. Each recording was assessed for audibility, completeness, adherence to the reference text, and correct assignment of the customer or agent role. Submissions containing excessive background noise, missing audio, incorrect content, or mismatches between the audio and reference transcription were rejected. Following quality control, the final dataset comprised 15,965 utterances totaling 30.63 hours of recorded speech.
\subsection{Audio Augmentation}
\label{sec:audio_augmentation}

To improve robustness to deployment conditions, we designed seven audio
augmentation pipelines based on an analysis of recordings collected over the
Ethio Telecom network. The pipelines model environmental noise,
recording-device imperfections, and telecommunication-channel artifacts. During training, augmentations were applied stochastically, with at most one telecommunication pipeline selected per utterance.

\subsubsection{Acoustic and Recording Augmentations}

\paragraph{Ambient room noise.}
We added a low-level stationary broadband noise floor and, with probability
$0.5$, applied sinusoidal amplitude modulation with a period of
$10s$. This pipeline approximates background noise and gradual
level variations caused by changes in microphone distance and speaker
orientation.

\begin{table}[t]
\centering
\scalebox{0.9}{
\begin{tabular}{lcc}
\toprule
\textbf{Source} & \textbf{Utterances} & \textbf{Hours} \\
\midrule
\multicolumn{3}{l}{\textit{Training data}} \\
WaxalNLP (amh)          & 38{,}022  & 189.0 \\
Waxal auto-labeled (amh)   & 80{,}321  & 398.1 \\
BDU Amharic ASR            & 32{,}901  & 101.0 \\
Common Voice (amh)         & 563       & 0.9   \\
Leyu -- Gonder dialect     & 11{,}339  & 112.8 \\
Leyu -- Wello dialect      & 10{,}010  & 114.7 \\
Leyu -- Gojjam dialect     & 10{,}095  & 95.8  \\
Leyu -- Shewa dialect      & 8{,}180   & 102.6 \\
\midrule
Train total                & 191{,}431 & 1{,}115.0 \\
\midrule
\multicolumn{3}{l}{\textit{Evaluation data (WaxalNLP)}} \\
Validation                 & 2{,}912   & 14.0  \\
\bottomrule
\end{tabular}
}
\caption{Composition of the combined Amharic ASR corpus for our model training. The training set is expanded by combining multiple publicly available Amharic ASR datasets. The validation set is limited to the WAXAL validation split, to which we apply our 0.4\% augmentations to provide a compact yet representative evaluation of the conditions covered by the test set.}

\label{tab:data-sources}
\end{table}

\paragraph{Distant / low-volume microphone.}
We attenuated the waveform uniformly by $15dB$ without introducing
noise or temporal distortions. This pipeline approximates reduced signal
levels associated with distant speakers and low-gain recording conditions.

\paragraph{Harsh environment.}
We applied a $12$-dB gain reduction, added stronger broadband noise, and
applied sinusoidal amplitude modulation with a period of $3s$.
This configuration models low signal levels, substantial background noise,
and rapid amplitude fluctuations under adverse recording conditions.

\subsubsection{Fixed Telecommunication Augmentations}

\paragraph{Wideband phone (VoLTE).}
We resampled the signal to $16kHz$, applied a
$200$–$3{,}500Hz$ band-pass filter, and performed dynamic-range
compression with a ratio of $2.5{:}1$. We then added low-level broadband
noise and stationary $50$-Hz sinusoidal interference to model channel noise
and power-line contamination.

\paragraph{Narrowband 2G + packet loss.}
We simulated a severely degraded narrowband channel by resampling the signal
to $8kHz$ and restricting its bandwidth to
$400$–$2{,}500Hz$. We applied dynamic-range compression with a
ratio of $6{:}1$, nonlinear \texttt{tanh}-based overdrive, heavy broadband
noise, and $50$-Hz interference. To approximate packet loss and transient
network disruptions, we inserted one to three complete amplitude dropouts
lasting $50$–$200ms$.

\subsubsection{Randomized Telecommunication Augmentations}

\paragraph{Randomized phone channel.}
To reduce overfitting to fixed degradation profiles, we introduced standard
and severe randomized variants with parameters sampled independently for
each utterance.

\textit{Standard variant.}
We used the standard telecommunication pipeline while sampling the lower and
upper band-pass cutoffs uniformly from $200$–$400Hz$ and
$3{,}000$–$4{,}000Hz$, respectively. We also randomized the
broadband noise amplitude, $50$-Hz interference amplitude, compression
threshold, and compression ratio.

\textit{Severe variant.}
We used the severely degraded narrowband pipeline, sampling the lower and
upper band-pass cutoffs from $300$–$600Hz$ and
$2{,}000$–$2{,}800Hz$, respectively. We randomized the
compression threshold and ratio, overdrive intensity, broadband noise
amplitude, and interference amplitude. In addition, we inserted one to five
dropouts lasting $20$–$400ms$ each. Each dropout either fully
suppressed the signal or attenuated it by a randomly selected factor.

\subsection{Training Data}

As shown in Table~\ref{tab:data-sources}, we combine several Amharic ASR datasets to investigate the effects of training-data scale and diversity on robustness tonoise and channel distortions. These incluede Google's WAXAL dataser \cite{diack2026waxallargescalemultilingualafrican}, automatically  kabled EthioASR data \cite{abdullah2026ethioasrjointmultilingualspeech}, BDU Speech \cite{Ejigu2025NoiseRobust}, the Amharic Common Voice ASR \footnote{\url{https://huggingface.co/datasets/Ibtihal02/amharic-commonvoice-asr}}, and Leyu Amahric dialectal dataset \footnote{\url{https://huggingface.co/leyu-amharic}}.



\section{Methodology}

\begin{figure*}
    \centering
\includegraphics[width= 0.9\textwidth]{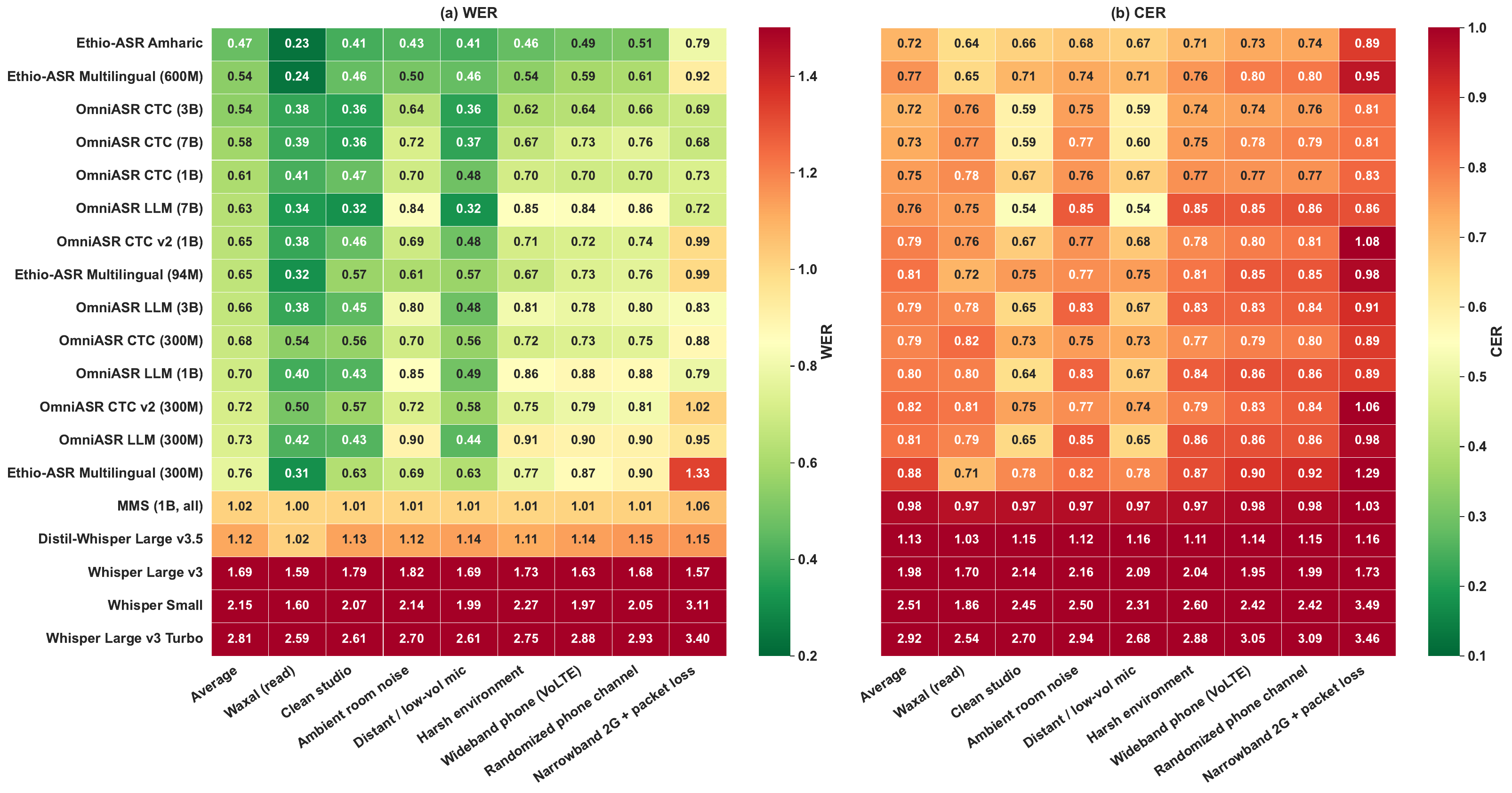}
    \caption{ASR performance across acoustic and telecommunication conditions. WER (left) and CER (right) are reported for all evaluated models across the average test set and individual degradation conditions, including weak signal, clean audio, ambient noise, distant/low-volume microphone, harsh environments, wideband phone (VoLTE), randomized phone channels, and narrowband 2G with packet loss. Lower values indicate better performance. The results highlight substantial variation in model robustness across realistic telecommunication conditions.}
    \label{fig:placeholder2}
\end{figure*}

To ensure the reproducible evaluation of the ASR models, we followed the approach adopted by the Open ASR Leaderboard~\cite{srivastav2025openasrleaderboardreproducible} and created configuration files for each dataset. All evaluations and model training were conducted using a single NVIDIA A100 GPU with 80~GB of memory.
\section{Result and Analysis}

For model training, we adopted the training framework provided by EthioASR~\cite{abdullah2026ethioasrjointmultilingualspeech}. We selected the best-performing model based on WER in our initial evaluation and continued fine-tuning from its pretrained checkpoint using our collected Amharic datasets. We extended the text-processing pipeline with number-to-text normalization by converting numerical digits into their corresponding Amharic word forms. During fine-tuning, audio augmentation was applied stochastically using Gaussian noise, random gain adjustments, and simulated environmental and telecommunication distortions, including amplitude modulation, band-pass filtering, codec resampling, dynamic-range compression, nonlinear distortion, power-line hum, and packet-loss dropouts. Following augmentation, the waveforms were peak-normalized when necessary to prevent clipping.

\paragraph{Evaluated Models.}

We evaluate a diverse set of ASR systems spanning Amharic-specific, multilingual, and general-purpose models. The evaluated systems include Ethio-ASR-Amharic and Ethio-ASR-multilingual models ~\cite{abdullah2026ethioasrjointmultilingualspeech} with 94M, 300M, and 600M parameters; OmniASR CTC models~\cite{omnilingualasr2025} with 300M, 1B, 3B, and 7B parameters; and OmniASR LLM models with 300M, 1B, 3B, and 7B parameters. We also evaluate the OmniASR-v2 300M and 1B CTC models, MMS-1B~\cite{pratap2023scaling}, Distil-Whisper~\cite{gandhi2023distilwhisperrobustknowledgedistillation}, and the Whisper family~\cite{radford2023robust}, including small, large-v3, and large-v3-turbo. Together, these models cover language-specific, multilingual, and general-purpose ASR approaches with substantially different model sizes and training paradigms.

\paragraph{Evaluation Normalization.}
Before computing WER and CER, we normalized both reference transcripts and model predictions using the same text-processing procedure. Numerical digits were converted to their corresponding Amharic word forms, punctuation was removed, and Amharic character variants were normalized. This ensures that the evaluation differences primarily reflect recognition errors rather than differences in formatting or character representation.


\textbf{RQ1:How robust are Amharic-specific, multilingual, and general-purpose ASR models to telecommunication-specific acoustic degradation?}

\label{sec:results}

We evaluated the robustness of the ASR systems across a range of acoustic and telecommunication conditions, including clean studio speech, environmental noise, low-energy speech, harsh acoustic environments, narrowband communication with packet loss, randomized telephone channels, and a wideband VoLTE-like channel. We report the word error rate (WER), where lower values indicate better recognition performance. The overall score reported in the average column is the arithmetic mean across the eight evaluation conditions. This provides a single measure of overall robustness while preserving the condition-specific results needed to identify sources of recognition degradation.

\paragraph{Overall Model Performance}

As shown in Figure~\ref{fig:placeholder2}, substantial differences in robustness can be observed across the evaluated model families. The Amharic-specific Ethio-ASR-Amharic model achieves the lowest overall WER of 46.57\%, making it the strongest-performing model across the complete evaluation suite. It is followed by the multilingual Ethio-ASR-Multilingual-600M model, with an average WER of 53.98\%. The best-performing general-purpose models are considerably less consistent across the different acoustic conditions.

The performance gap between the two largest Amharic-oriented models is particularly informative. Ethio-ASR-Amharic improves over Ethio-ASR-Multilingual-600M, reducing the average WER from 53.98\% to 46.57\%, corresponding to a relative reduction of approximately 13.7\%. This result suggests that increasing multilingual coverage alone does not necessarily provide the same benefit as adapting acoustic and linguistic representations to the target language and deployment domain. This finding is especially relevant for low-resource ASR, where a model can benefit substantially from specialization, even when its parameter count is comparable to that of a multilingual alternative.

The results also demonstrate that model size is not a sufficient predictor of robustness. Within the OmniASR family, increasing the parameter count does not consistently lead to lower WER. For example, OmniASR-CTC-7B obtains an average WER of 58.32\%, whereas the smaller OmniASR-CTC-1B and OmniASR-CTC-300M models obtain 61.07\% and 67.94\%, respectively. Although the larger model performs better than these smaller variants overall, the relationship is not monotonic across all model families and conditions, despite their relatively large parameter counts. These observations indicate that robustness to telecommunication distortion depends not only on model capacity but also on language specialization, training distribution, architecture, and exposure to channel variability.

\paragraph{Performance Under Clean and Acoustic Conditions}

Under relatively clean conditions, Ethio-ASR-Amharic achieved the best WAXAL WER (23.25\%), closely followed by Ethio-ASR-Multilingual-600M (23.82\%). However, OmniASR-LLM-7B performed best on clean studio speech (31.59\%), while Ethio-ASR-Amharic achieved 40.5\%, showing that the Amharic-specific model is not uniformly superior under clean conditions.

Under ambient room noise, Ethio-ASR-Amharic achieved 42.74\%, substantially outperforming the OmniASR models (64.18--89.70\%), indicating stronger resilience to environmental noise. However, for distant/low-volume speech, OmniASR-LLM-7B achieved the lowest WER (31.81\%), compared with 40.59\% for Ethio-ASR-Amharic. Overall, these results show that model strengths vary by acoustic condition, while the Amharic-specific model provides greater consistency across conditions.

\textbf{RQ2: How does ASR performance change as the severity and combination of telecommunication degradations increase?}

\begin{figure*}
    \centering
    \includegraphics[width=0.9\textwidth]{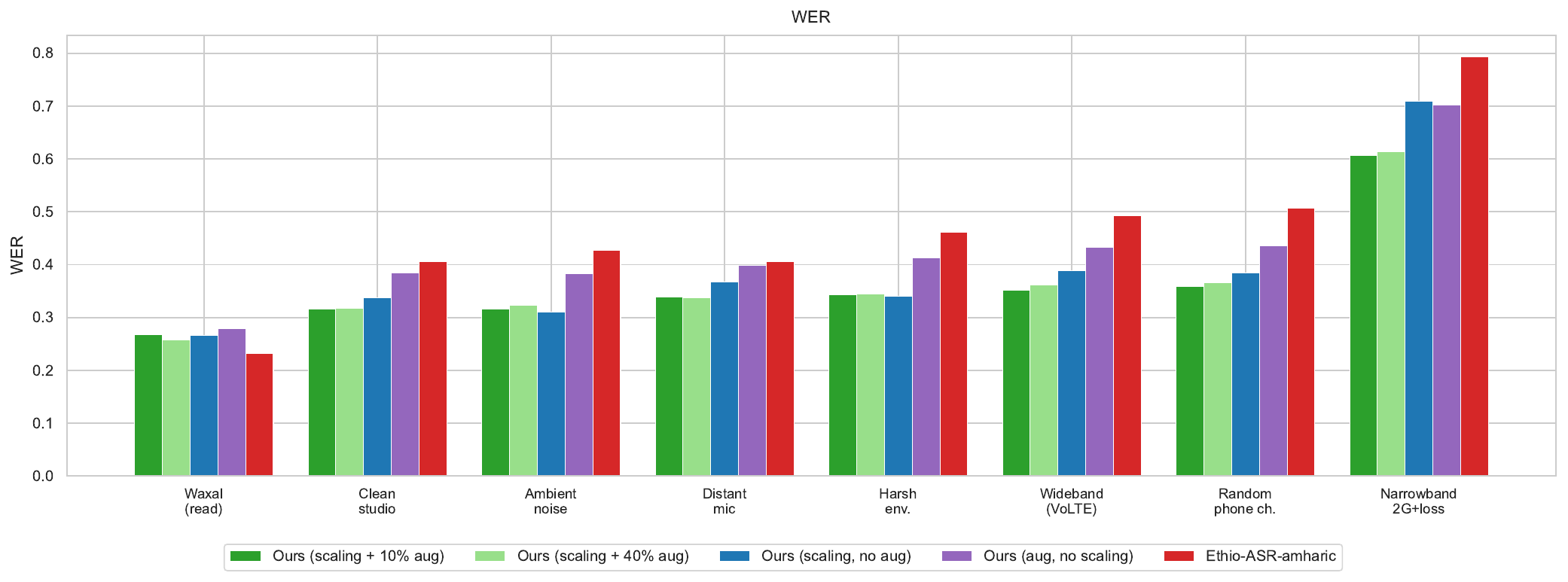}
    \caption{Word Error Rate (WER, lower is better) of our fine-tuned Amharic ASR
  models compared with the strongest publicly available baseline
  (Ethio-ASR-amharic) across eight evaluation conditions, ordered from least to
  most acoustically degraded.
  }
    \label{fig:finetuning}
\end{figure*}

The largest performance degradation occurs under telecommunication conditions, particularly when narrowband communication is combined with packet loss. For Ethio-ASR-Amharic, WER increases from 40.57\% under clean studio conditions to 79.36\% under narrowband 2G with packet loss. This corresponds to an absolute increase of 38.79 percentage points and a relative increase of 95.6\%. This substantial degradation demonstrates that the recognition performance under clean speech is not a reliable indicator of performance under realistic telecommunication conditions.

The narrowband 2G condition is the most challenging condition for most evaluated models. Ethio-ASR-Multilingual-600M reaches a WER of 92.13\%, while Ethio-ASR-Multilingual-94M and Ethio-ASR-Multilingual-300M reach 99.39\% and 132.75\%, respectively. The deterioration is therefore particularly pronounced for the multilingual Ethio-ASR variants. In contrast, OmniASR-CTC-7B achieves a WER of 68.03\%, the best result under this condition, closely followed by OmniASR-CTC-3B at 68.52\%.


The randomized phone-channel condition is particularly informative because it evaluates robustness to a distribution of channel characteristics rather than a single fixed degradation profile. Ethio-ASR-Amharic achieves the lowest WER at 50.67\%, followed by Ethio-ASR-Multilingual-600M at 60.78\%. This result supports the use of randomized channel conditions for robustness evaluation, as performance on a fixed channel does not necessarily predict robustness to broader channel variability.

A similar trend is observed for the wideband phone (VoLTE) condition. Ethio-ASR-Amharic achieves a WER of 49.29\%, compared with 59.35\% for Ethio-ASR-Multilingual-600M, giving the Amharic-specific model an advantage of approximately 10.06 percentage points. The consistent advantage across randomized and standardized telephone conditions further highlights the robustness of the Amharic-specific model.

Interestingly, the performance gap between clean and telephone conditions varies substantially across model families. For Ethio-ASR-Amharic, WER increases from 40.57\% under clean studio speech to 49.29\% under wideband telephone conditions. In contrast, OmniASR-LLM-7B increases from 31.59\% to 84.33\%, representing a much larger degradation despite its stronger clean-speech performance. These results highlight the limitations of evaluating ASR systems exclusively on clean benchmarks: strong clean-speech accuracy does not necessarily translate into robustness under realistic channel transformations.

\textbf{RQ3: Can data scaling and targeted augmentation improve the robustness of Amharic ASR models under telecommunication conditions?}

Figure~\ref{fig:finetuning} compares fine-tuning variants across WAXAL read speech, clean studio speech, ambient room noise, distant/low-volume microphone recordings, harsh environments, wideband telephone (VoLTE), randomized telephone channels, and narrowband 2G with packet loss. We ablate two training factors: data scaling, using the full training corpus, and simulated telecommunication-channel augmentation applied to either 10\% or 40\% of the training utterances.

The combination of data scaling and 10\% augmentation achieves the lowest WER across the degraded evaluation conditions. For example, under narrowband 2G with packet loss, for example, it reduces the WER from 0.794 for the baseline to 0.607, corresponding to a 23.6\% relative WER reduction. Data scaling alone consistently outperforms augmentation alone, suggesting that increased training data coverage provides the largest contribution to robustness, while targeted augmentation provides complementary gains, particularly under more severe channel conditions.

Interestingly, the baseline performs best only on the in-domain WAXAL read-speech condition, achieving a WER of 0.232 compared with 0.258 for the best fine-tuned variant. This trade-off suggests that fine-tuning with larger and more diverse data can improve robustness to deployment conditions while causing a small reduction in performance on the original in-domain distribution.

\section{Conclusion and Future Work}

In this work, we investigated the applicability and robustness of ASR models for low-resource languages, focusing on Amharic in the context of telecommunications. Existing Amharic ASR datasets provide limited coverage of domain-specific and noisy acoustic conditions, making it difficult to assess model robustness in real-world telecom scenarios. To address this gap, we introduced TeleASR-Amharic, a 30-hour benchmark covering seven diverse audio conditions designed to reflect realistic telecommunication environments. Our evaluation highlights the key challenges in deploying ASR models under acoustic and channel degradation and demonstrates that both data scaling and domain-specific augmentation can substantially improve ASR robustness.

As a next step, we plan to extend this conversational ASR resource to the evaluation of speech-to-speech systems. In particular, the dataset can provide a foundation for studying whether improvements in speech recognition under realistic conversational and telecommunication conditions translate into more reliable end-to-end spoken language systems.

\section{Ethical Consideration}
This study addresses privacy , fairness, and responsible use in developing robust Amharic ASR for telecommunication speech.


\paragraph{Privacy and Data Governance.} Speech data are handled according to applicable data-use agreements, institutional requirements, and privacy regulations. Identifying metadata are protected where applicable, access to non-public recordings is restricted, and raw recordings are not publicly released.


\paragraph{Synthetic Data and Benchmark Construction.}
Synthetic augmentation diversifies acoustic realizations but does not remove source-data biases in speaker, language variety, or recording-condition representation. 



\paragraph{Speaker Representation and Fairness.} Recognition performance is compared across the  available male and female speaker groups. The differences are interpreted as model performance variations and not inherent speaker differences. The findings are limited by the demographic coverage of the  dataset  and cannot be generalized beyond the evaluated population.


\paragraph{Responsible Use.} ASR systems may be used in sensitive or high-stakes settings; therefore, the results do not establish suitability for unrestricted legal, medical, law enforcement, or surveillance applications. Performance degradation under certain telecommunications conditions highlights the need for deployment-specific validation.


\paragraph{Responsible Release and Reproducibility.} Reproducibility is supported through benchmark descriptions, preprocessing procedures, augmentation configurations, evaluation scripts, and other non-sensitive resources when raw recordings cannot be shared. Synthetic degradations are distinguished from observed telecommunications artifacts to avoid overstating their real-world correspondence.


Overall, this study promotes robust Amharic ASR through privacy, fairness, reproducibility, and responsible deployment within the study's data and acoustic conditions.



\section*{Limitations}

TeleASR-Amharic is currently limited to evaluating the robustness of ASR under
telecommunication conditions. However, the small pool of speakers may not fully represent the variations in the different dialects of Amharic. The lack of publicly available real-call audio also limits validation under real-world conditions, including authentic mobile codec artifacts and other channel effects. Privacy and consent requirements may further restrict access to real-call recordings. In addition, data augmentation may not capture all real-world variations, potentially limiting generalization to unseen conditions. Future work can address these limitations by including more diverse speakers, validating simulated conditions against  real-call data where feasible, and extending the benchmark to evaluate end-to-end speech systems and other components of spoken telecommunications.


\section*{Acknowledgments}
This research was supported by the GSMA  through the \href{https://www.gsma.com/get-involved/gsma-foundry/the-african-ai-languages-model-project/}{African AI Language Models project}. The views and opinions expressed in this publication are those of the authors and do not necessarily reflect the views or positions of the GSMA.

\bibliography{custom}

\appendix

\label{sec:appendix}
\section{Error analysis Visualization}
\begin{figure}[ht]
    \centering
    \caption{Error analysis platform developed to identify failure patterns in the dataset and inform evaluation gaps. }\includegraphics[width=0.9\linewidth]{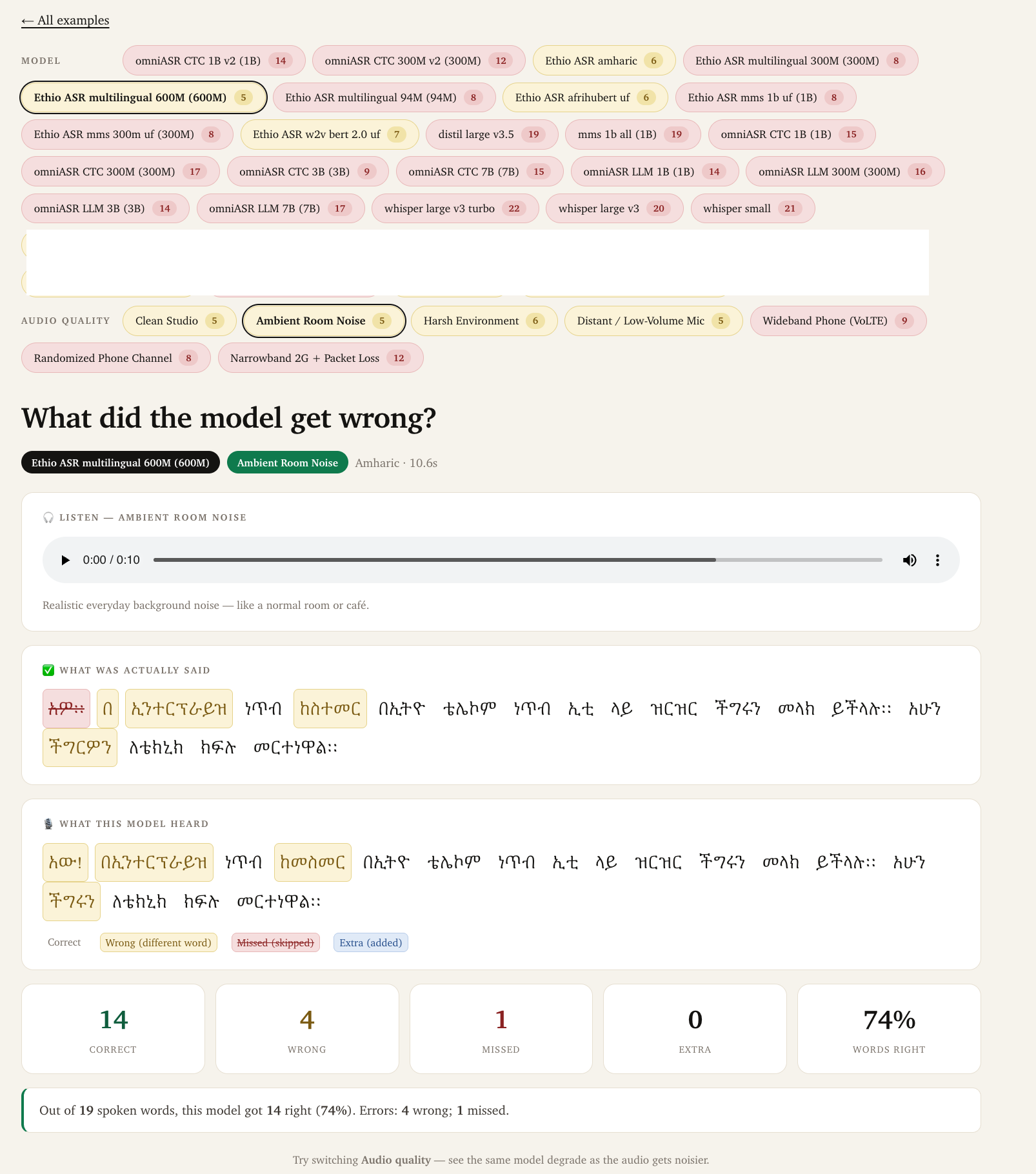}
    \label{fig:error}
\end{figure}

In this example shown in Figure~\ref{fig:error}, the models perform correctly, while the apparent errors arise from metric computation, highlighting the need for punctuation removal. Text normalization and prefix normalization can further improve evaluation by distinguishing metric-related discrepancies from genuine model failures. Some models hidden for anonymity.

\section{Audio Noise variation Analysis}
\begin{figure}[ht]
    \centering
    \caption{Evaluation of Generated Augmentations. We consulted telecommunication experts to assess whether the generated augmentations accurately reflected real-world telecommunication noise and channel conditions.}
    \label{fig:signal}
    \includegraphics[width=\linewidth]{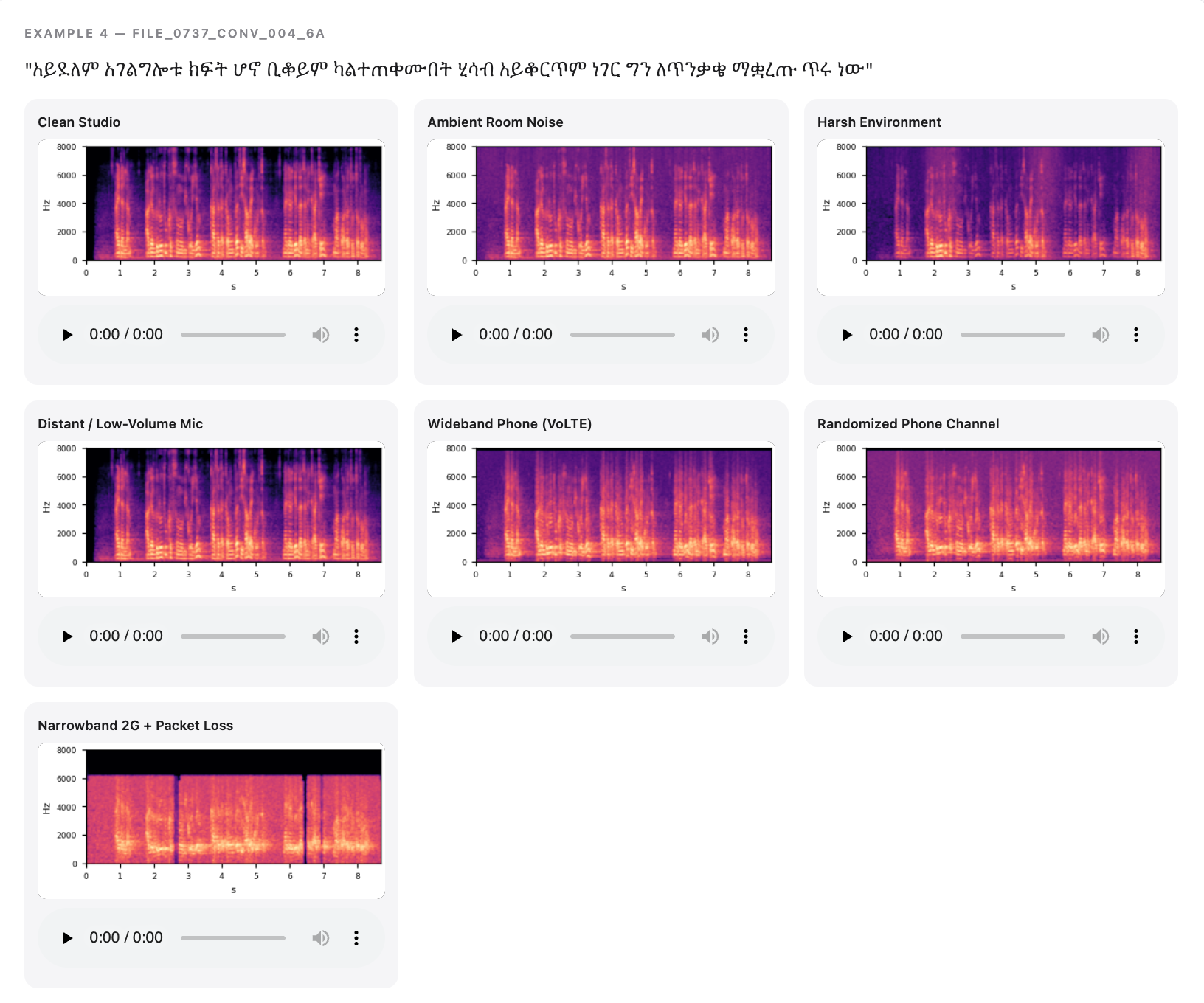}
\end{figure}

 In Figure~\ref{fig:signal}, we applied the same expert-driven evaluation procedure to real-world telecommunication audio recordings, using them to identify and characterize the signal variations observed in practical communication channels. These observations subsequently guided the design and calibration of the simulated augmentations, ensuring that the resulting conditions captured the acoustic and signal-level characteristics of real telecommunication audio.
 
\end{document}